# Enhancing Construction Site Analysis and Understanding with 3D Segmentation

Sri Ramana Saketh Vasanthawada[1]; Pengkun Liu[2]; and Pingbo Tang, Ph.D.[3]

[1] Civil and Environmental Engineering, Carnegie Mellon University, 5000 Forbes Avenue, Pittsburgh, PA 15213. Email: vsrirama@andrew.cmu.edu
[2] Civil and Environmental Engineering, Carnegie Mellon University, 5000 Forbes Avenue, Pittsburgh, PA 15213; Email: pengkunl@andrew.cmu.edu
[3] Associate Professor, Civil, and Environmental Engineering, Carnegie Mellon University, 5000 Forbes Avenue, Pittsburgh, PA 15213 (corresponding author). Email: ptang@andrew.cmu.edu

**ABSTRACT**

Monitoring construction progress is crucial yet resource-intensive, prompting the exploration of computer-vision-based methodologies for enhanced efficiency and scalability. Traditional data acquisition methods, primarily focusing on indoor environments, falter in construction site's complex, cluttered, and dynamically changing conditions. This paper critically evaluates the application of two advanced 3D segmentation methods, Segment Anything Model (SAM) and Mask3D, in challenging outdoor and indoor conditions. Trained initially on indoor datasets, both models' adaptability and performance are assessed in real-world construction settings, highlighting the gap in current segmentation approaches due to the absence of benchmarks for outdoor scenarios. Through a comparative analysis, this study not only showcases the relative effectiveness of SAM and Mask3D but also addresses the critical need for tailored segmentation workflows capable of extracting actionable insights from construction site data, thereby advancing the field towards more automated and precise monitoring techniques.

**INTRODUCTION**

Construction progress monitoring accounts for significant time and resources in a project's life cycle. Traditionally, progress monitoring is performed using labor, safety logs, and photographs. Since the rise of accessible hardware such as laser scanners, various stakeholders in a construction project typically prefer periodically capturing data with the goal to gain site understanding and improve efficiency. Despite point cloud being a rich data source, and providing abundant information, the industry lacks accessible and robust robustly adaptable progress monitoring mechanisms or workflows, particularly catering to different scenes with various geometric and lighting conditions. This is because the construction site has a dynamically changing environment with various moving parts. Therefore, current techniques impede gaining adequate spatial awareness to stakeholders to identify workflow bottlenecks. Identifying bottlenecks can improve adherence to schedule, optimization of resources, and risk management. Considering the benefits, researchers are making active strides to develop accessible and advanced technologies to understand the spatial interrelationship between elements on-site better to generate actionable insights. Such efforts increased the application of advanced and computational data acquisition techniques for project performance enhancement.

Over the years, researchers have extensively worked on integrating point clouds and computational approaches due to their potential to deliver meaningful insights (Grilli et al., 2021; Pal et al., 2023). However, these approaches were limited to completed structures or stable indoor environments rather than active construction sites with higher occlusions or varying lighting conditions. The lack of relevant research can be attributed to the unavailability of datasets. For instance, researchers have utilized PointNet and other related architectures to successfully segment common indoor objects like doors, ceilings, floors, and windows (Liang et al., 2021; Zhao et al., 2020) due to the Stanford 3D Indoor Scene Dataset (S3DIS) (Armeni et al., 2017). However, acquiring and labeling every dataset is cost and time-prohibitive, considering the complexity and volume of classes. S3DIS involved 6-large scale indoor areas with 271 rooms. Labeling such a high volume of images is cumbersome. Despite having success in indoor segmentation, the same cannot be translated into an active construction scene. Construction sites are highly dynamic. Therefore, it is important for an approach that is automated and can recognize scenarios for effectively gaining spatial awareness of the site.

Identifying this challenge, this research explores the efficacy of 3D segmentation methods in cluttered indoor and outdoor site conditions for construction site understanding. It includes the two state-of-the-art methods, Segment Anything Model (SAM) and Mask 3D, two transformer-based frameworks on an active construction site to understand their effectiveness and bottlenecks under various conditions. This effort allows future researchers to weigh the merits and demerits of each model under varying scenarios where the user is expected to have varying conditions. SAM is a transformer-based foundational model for 2D image segmentation and object identification. In the current study, we translated the 2D SAM to a 3D context, allowing to segment point clouds. On the other hand, Mask3D is also a transformer-based model with a sparse convolutional feature backbone with query refinement. The model has been examined on multiple indoor datasets. Therefore, this research answers two research questions (RQ):

*RQ 1: To what extent can the current state-of-the-art methods, SAM and Mask 3D, effectively be used to gain a scene understanding of a complex construction site?*
*RQ 2: How competitive is SAM compared to other pre-trained models in a similar context regarding segmentation accuracy and adaptation to high-occlusion and lighting-varying construction site environments?*

**METHODS**

This study employs a novel approach to understand construction site activities, leveraging the capabilities of 3D laser scanning technology as depicted in Figure 1. Our methodology is twofold, involving data preparation and analysis through two distinct computational models: SAM and Mask3D. Each model plays a pivotal role in our framework, aimed at optimizing the segmentation and understanding of construction site dynamics.

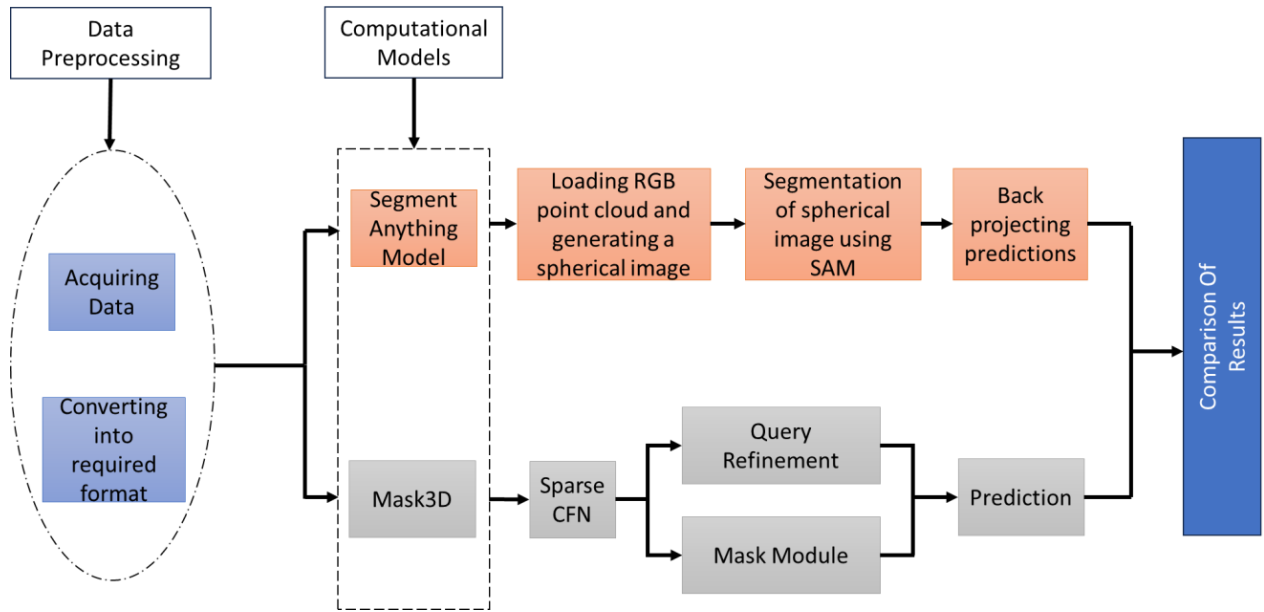

**Figure 1. The framework for construction site understanding through 3D laser scanning.**

**SAM Model Implementation.** The SAM model is structured into three layers: an image encoder, a prompt encoder, and a mask decoder. Initially, an image with a prompt is provided to SAM. The Vision Transformer (ViT) based encoder generates image embeddings, and the prompt encoder converts the prompts into vector embeddings. These two sets of embeddings are combined and processed by the mask decoder, which predicts masks based on the combination. However, existing literature on 3D segmentation lacks a definitive workflow, prompting this study to adopt a new approach by (Grilli et al., 2021), integrating the Segment Anything Model (SAM) into our workflow. Our process begins with data preparation in standard point cloud formats and proceeds to generate a spherical image from the RGB point cloud, maintaining a pixel-to-point mapping essential for 3D segmentation. SAM facilitates unsupervised segmentation, applying masks based on the point cloud's center. Subsequent segmentation allows back-projection at the point level, culminating in a fully segmented point cloud.

**Mask3D Model Implementation.** Mask3D also adopts a transformer-based architecture that processes instance masks in parallel, leveraging the transformer's learning mechanism. Mask3D approach by (Schult et al., 2023) provides a user interface that allows for class prediction based on the ScanNet which consists of different classes, including a set of indoor objects.

**DATA FOR CONSTRUCTION SITE SEGMENTATION AND UNDERSTANDING**

The dataset critical to this research encompasses temporal point cloud data from Scaife Hall, a newly constructed academic building at Carnegie Mellon University. Utilizing FARO S350+ and S120 laser scanners, we compiled over 40 distinct point clouds, showcasing a broad spectrum of construction activities and equipment variations. This rich dataset is instrumental in delineating the contrasts in segmentation capabilities between the Segment Anything Model (SAM) and Mask3D.

**Data Acquisition:** The acquisition process, leveraging advanced laser scanning technology, ensured the collection of high-density point clouds. This method allowed for capturing the intricate details of construction sites, including varying equipment types and activity levels, essential for evaluating the models' performance under realistic conditions.

**Data Preparation:** Before analysis, the point clouds underwent a meticulous preparation phase. This data preparation included aligning the scans to a unified coordinate system and cleaning to remove any irrelevant objects or noise that could skew the segmentation process. The preparation ensured that the input data for both SAM and Mask3D was of acceptable quality, facilitating a fair comparison of their segmentation prowess.

**Dataset Characteristics:** The chosen dataset is particularly notable for its diversity in construction site conditions, ranging from densely packed areas with significant occlusions to more open spaces with varied lighting. This variability is crucial for assessing the models' adaptability and accuracy across various real-world scenarios. Additionally, the temporal aspect of the dataset provides insight into the dynamic nature of construction sites, offering a comprehensive understanding of how segmentation models can track changes over time.

**RESULTS AND DISCUSSIONS**

Our study conducted an in-depth analysis utilizing the point cloud data from Scaife Hall, with a particular focus on Scan 035 captured at the building's second level (Figure 2). The picture in the left of Figure 2 is spherical image obtained from the point cloud and the right contains the segmentation using SAM. In the current example, the scan is rich with construction elements such as pipes, boxes, and scaffolds, serving as a basis for evaluating the efficacy of our segmentation methodology.

**SAM Model Performance:** The segmentation results, as visually represented in Figure 3, highlight SAM's proficiency in distinguishing common construction objects within the cluttered scene. For instance, the box on the right has a clear division between the areas, showing the potential to segment objects more accurately. Similarly, the other parts of the image show that the objects were accurately segmented based on visual assessment. Using this approach, multiple point clouds were segmented to assess whether SAM 3D can be generalizable in most scenarios. The 2D SAM images were backprojected to the point cloud. Various transformation were performed to obtain compute the back-projection from a 2D image to point cloud. This includes resulting in point cloud segmentation. This way, the point cloud segmentation has been performed, with good results, when assessed visually.

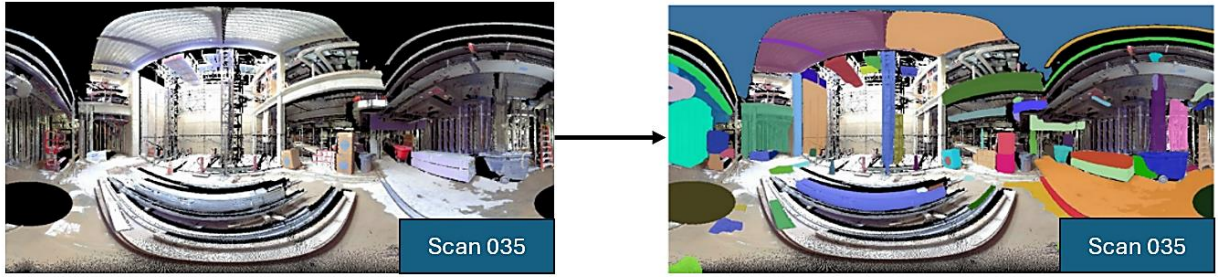

**Figure 2: Panoramic Image of the scanning location and segmentation using SAM**

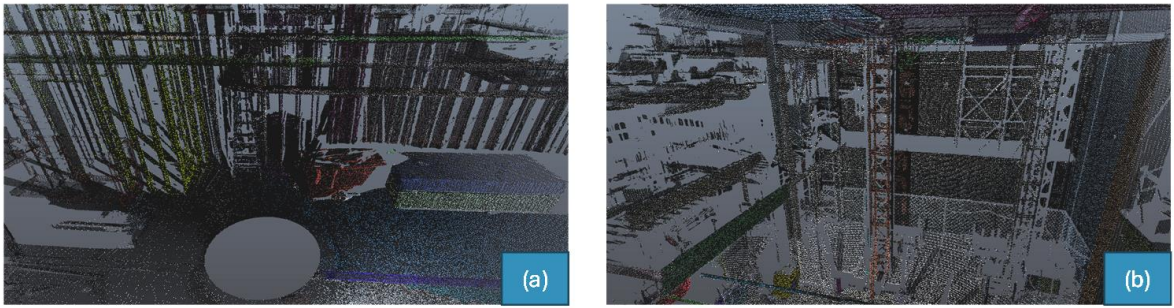

**Figure 3: Back projected and segmented point cloud from 2D masks with SAM**

**Mask3D Model Performance:** To understand the variation with SAM, Mask3D has been tested, and the results are as follows. Visual inspection of Figure 4 shows that the classification of the objects is not accurate, and segmentation is accurate. This result is because the objects trained by Mask3D are trained in the standard indoor scenes, not construction field operation scenes. On the other hand, objects which were not included in the training set like columns and pipes were broken down into smaller parts and identified as different objects.

For example, although the scaffold has been segmented in Figure 4 (a), the object has been labeled as a shower curtain. Such misclassifications can be retraced to training datasets and here, SAM outperforms Mask 3D. SAM was able to identify the whole scaffold as a singular object. Similarly, overhead pipes in Figure 4 (b) are not uniform and have been broken down into multiple parts, but SAM was able to identify all the objects as a singular entity.

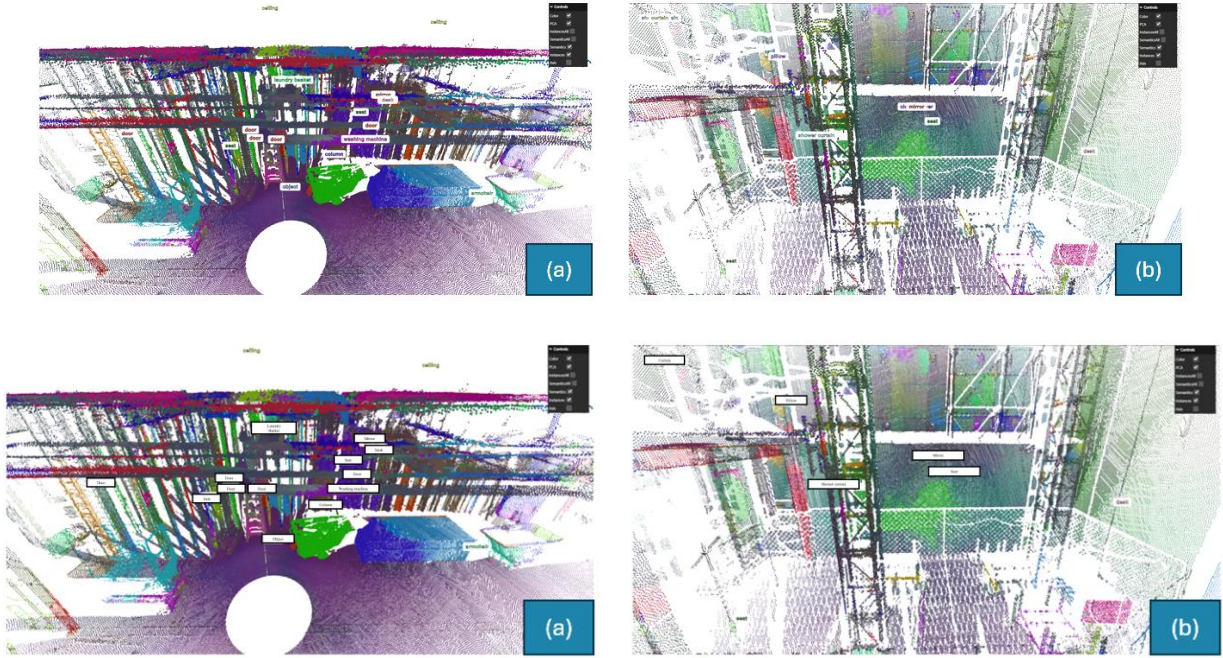

**Figure 4: Front (a) and Rear (b) view of the processed point cloud with Mask3D.**

To understand instances where each model would work effectively, the study also analyzed datasets experiencing different lighting and objects within the scene. The subsequent Table 1 provides an overview of the results. From Table 1, it is evident that SAM and Mask3D excel at their tasks of segmentation and classification. However, the accuracy and consistency of the models largely vary. For instance, the accuracy of SAM is dependent on the method of back projecting the points. The model accurately segments the subjects when they are in the 2D stage due to its robustness with images. However, the back projection of points in the 3D stage falls short in a few instances where far-away objects are not segmented. This is because SAM only segments the subjects that are visible in the 2D images. The cluttered sections in the images are not identified and therefore are completely ignored in the analysis. This is one of the significant limitations of SAM. Future research should be directed towards improving back projection techniques which can improve the understanding of a construction scene. On the other hand, Mask3D tries classifying all the point clouds. As the training data for Mask3D comprises indoor objects, the objects are mostly misclassified. However, it is to be noted that Mask3D tries to process every data point. Translating this key feature into SAM can be instrumental as this can provide a comprehensive understanding of the point cloud. However, a cumbersome process, a potential method to improve point cloud classification is by creating a curated dataset.

**CONCLUSION**

The comparative analysis between SAM and Mask3D underscores the nuanced capabilities and constraints of each model in the context of construction site segmentation. SAM's strength lies in its robust image-based segmentation, demonstrating significant potential for accurate delineation of site elements. However, its performance is contingent upon the clarity and proximity of objects, revealing limitations in scenarios with complex occlusions or distant objects. Conversely, Mask3D's broader attempt at classifying all observed elements, despite its misclassification issues, highlights the importance of comprehensive training datasets. Its methodology suggests an avenue for enhancing segmentation through extensive data processing, although it necessitates refinement in object recognition and classification accuracy.

These findings provoke consideration of hybrid approaches that amalgamate SAM's image-based precision with Mask3D's exhaustive data processing capabilities. Future studies could explore the development of models that leverage the strengths of both, potentially through enriched training datasets or advanced back-projection techniques to ameliorate the identified limitations.

Table 1: Comparative observations between SAM and Mask3D.

| RGB Image | SAM | Mask3D | Observations |
| --- | --- | --- | --- |
| 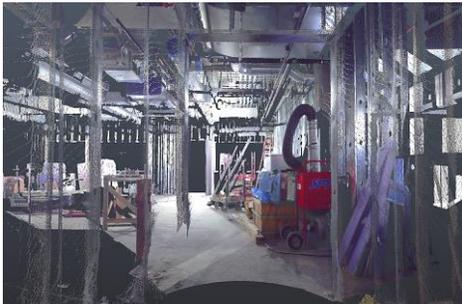 | 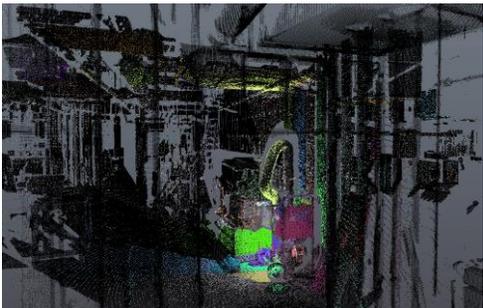 | 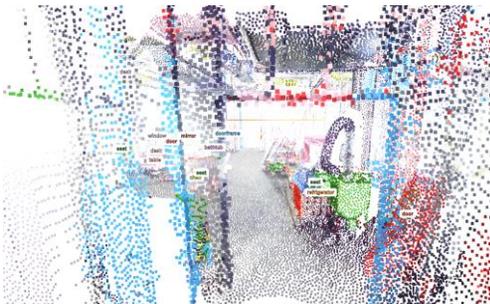 | SAM was not able to segment cluttered and closely related objects, where Mask3D assigned a class to the cluttered steel frames. |
| 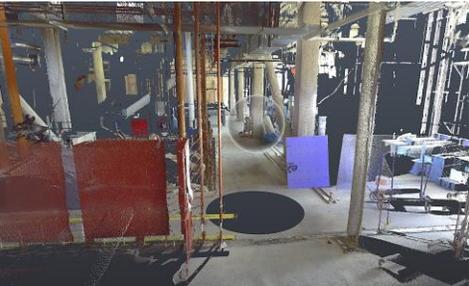 | 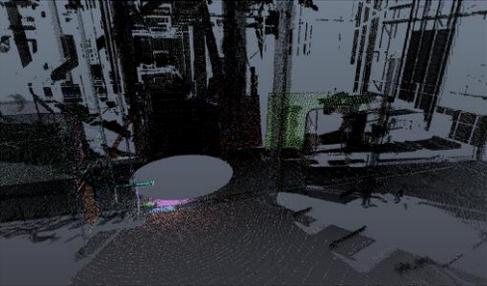 | 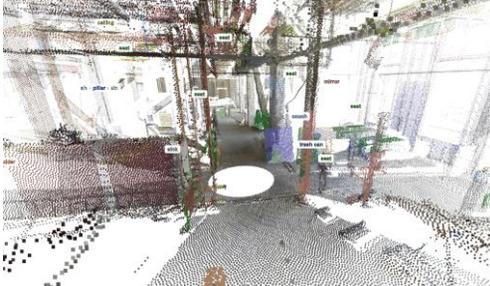 | SAM struggles to identify closely clustered objects. At the same time, Mask3D assigned a class to the objects. |
| 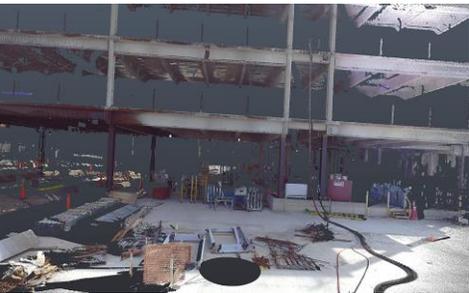 | 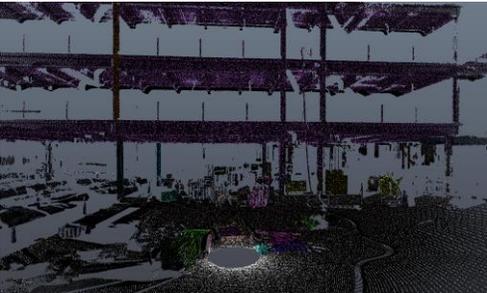 | 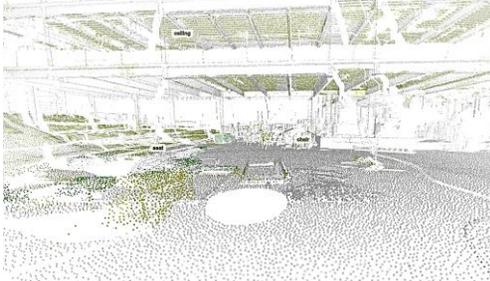 | SAM was able to segment construction equipment and steel frame of the building accurately as compared to Mask3D |

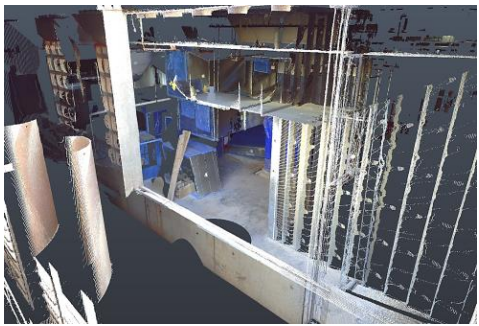 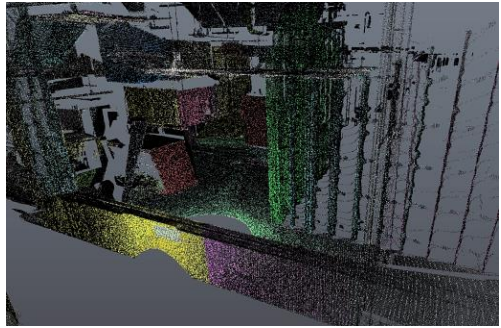 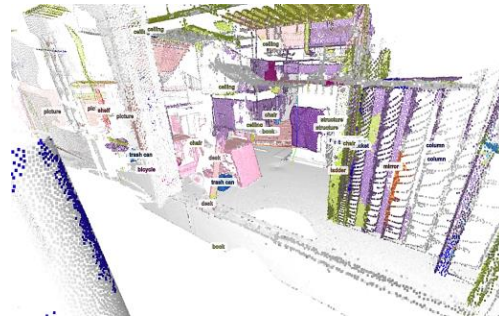

Mask3D and SAM yield exceptional and similar results primarily due to better lighting conditions and objects not being close together.

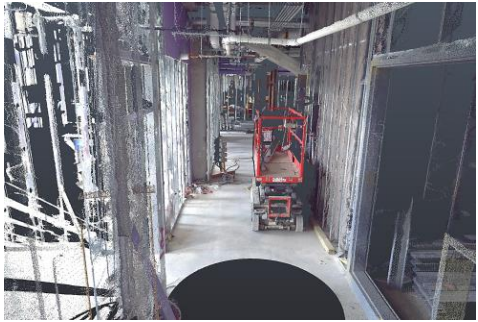 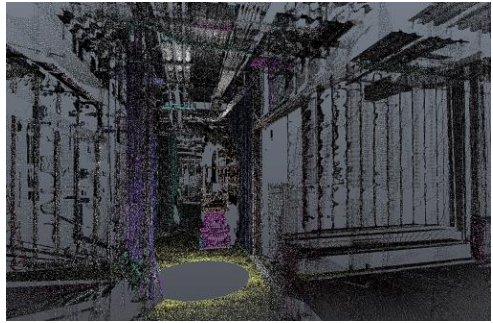 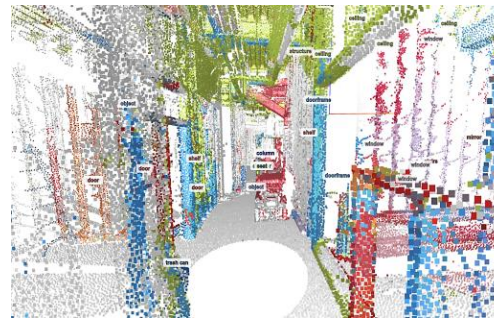

SAM and Mask3D yield very similar results, however, SAM could segment the overhead pipes and ducts individually rather than Mask3D's grouping.

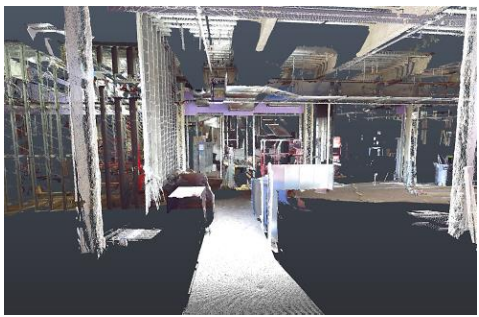 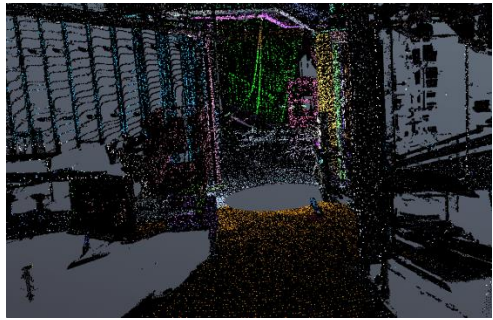 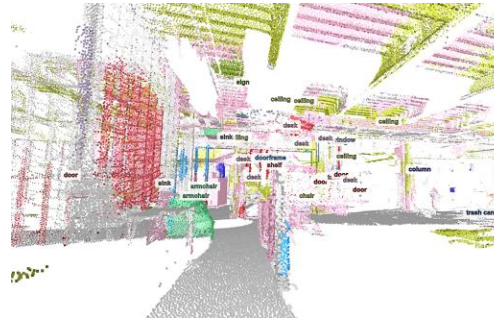

Mask3D resulted in better segmentation and classification of pipes and other equipment as compared to SAM.

**ACKNOWLEDGEMENTS**

The research presented in this paper was carried out with the support of the Summer Research Program of the Department of Civil and Environmental Engineering at Carnegie Mellon University. The authors express their gratitude to Dr. Yujie Wei and Zichen Wang for generously providing the point cloud data that was foundational in conducting this research.**ACKNOWLEDGEMENTS**

The research presented in this paper was carried out with the support of the Summer Research Program of the Department of Civil and Environmental Engineering at Carnegie Mellon University. The authors express their gratitude to Dr. Yujie Wei and Zichen Wang for generously providing the point cloud data that was foundational in conducting this research.